\documentclass{article}

\usepackage[preprint]{neurips_2025}

\usepackage{bm}
\usepackage{ragged2e} 
\usepackage{booktabs,makecell, multirow, tabularx}
\usepackage{subcaption}
\usepackage{multirow} 
\usepackage{colortbl}
\usepackage{color}
\usepackage{enumitem}
\usepackage{multirow}
\usepackage{graphicx}
\usepackage{wrapfig}
\usepackage{amsmath}
\usepackage{pifont}
\usepackage{caption}
\usepackage{wrapfig}
\usepackage{amsmath}
\usepackage{colortbl}
\usepackage{color}
\usepackage{enumitem}
\usepackage{amssymb} 
\usepackage[utf8]{inputenc} 
\usepackage[T1]{fontenc}    
\usepackage{hyperref}       
\usepackage{url}            
\usepackage{booktabs}       
\usepackage{amsfonts}       
\usepackage{nicefrac}       
\usepackage{microtype}      
\usepackage{xcolor}         
\usepackage{graphicx}
\usepackage{multirow}
\usepackage{xcolor}
\usepackage{caption}
\definecolor{cvprblue}{rgb}{0.21,0.49,0.74}
\newcommand{\methodshort}[1]{\textsc{Sparse-vDiT}}
\usepackage{algorithm}
\usepackage{algorithmic}
\usepackage[algo2e, ruled]{algorithm2e}
\usepackage{graphicx}
\usepackage{wrapfig}
\usepackage{mathrsfs}
\usepackage{amssymb}
\SetKwComment{Comment}{$\triangleright$\ }{}

\hypersetup{
    colorlinks=true,
    linkcolor=red,
    citecolor=cyan,
    filecolor=magenta,      
    urlcolor=magenta,
}

\title{Sparse-vDiT: Unleashing the Power of Sparse Attention to Accelerate Video Diffusion Transformers}

\author{
Pengtao Chen$^1$ \quad Xianfang Zeng$^{2 \text{\ddag}}$ \quad Maosen Zhao$^1$ \quad Peng Ye$^3$  \\
\textbf{Mingzhu Shen}$^4$ \quad \textbf{Wei Cheng}$^2$ \quad \textbf{Gang Yu}$^2$ \quad \textbf{Tao Chen}$^{1}$\thanks{\noindent Corresponding author. $^{\ddag}$Project leader. Work was done when interned at StepFun.  \vspace{-15pt}} \\
$^{1}$ Fudan University \quad $^{2}$ StepFun \quad $^{3}$ The Chinese University of Hong Kong\\
$^{4}$ Imperial College London\\
{\texttt{Code:} \href{https://github.com/Peyton-Chen/Sparse-vDiT}{\textbf{\texttt{https://github.com/Peyton-Chen/Sparse-vDiT}}}}
}

\begin{document}

\maketitle

\begin{abstract}
While Diffusion Transformers (DiTs) have achieved breakthroughs in video generation, this long sequence generation task remains constrained by the quadratic complexity of attention mechanisms, resulting in significant inference latency. Through detailed analysis of attention maps in Video Diffusion Transformer (vDiT), we identify three recurring sparsity patterns: diagonal, multi-diagonal, and vertical-stripe structures.   And even 3-6\% attention heads can be skipped. Crucially, these patterns exhibit strong layer-depth and head-position correlations but show limited dependence on the input content. Leveraging these findings, we propose Sparse-vDiT, a sparsity acceleration framework for vDiT comprising: 1) Pattern-optimized sparse kernels that replace dense attention with computationally efficient implementations for each identified sparsity pattern. 2) An offline sparse diffusion search algorithm that selects the optimal sparse computation strategy per layer and head via hardware-aware cost modeling. After determining the optimal configuration, we fuse heads within the same layer that share the same attention strategy, enhancing inference efficiency. Integrated into state-of-the-art vDiT models (CogVideoX1.5, HunyuanVideo, and Wan2.1), Sparse-vDiT achieves 2.09$\times$, 2.38$\times$, and 1.67$\times$ theoretical FLOP reduction, and actual inference speedups of 1.76$\times$, 1.85$\times$, and 1.58$\times$, respectively, while maintaining high visual fidelity, with PSNR values reaching 24.13, 27.09, and 22.59. Our work demonstrates that latent structural sparsity in vDiTs can be systematically exploited for long video synthesis. 
\end{abstract}

\section{Introduction}

In recent years, diffusion models have achieved significant advances in image generation~\cite{rombach2022high}, prompting growing interest in extending them to video synthesis. Early approaches, such as SVD~\cite{SVD2023blattmann} and Dynamicrafter~\cite{dynamicrafter2024xing}, employed a 2D+1D framework that provided computational efficiency but lacked real-time interaction between spatial and temporal features, resulting in limited spatiotemporal consistency. Recent progress in 3D full-attention Video Diffusion Transformers (vDiT)~\cite{peebles2023DiT} has effectively addressed these limitations. Built on this foundation, models such as OpenSora~\cite{opensora2024lin}, CogVideoX~\cite{cogvideox2024yang}, HunyuanVideo~\cite{kong2024hunyuanvideo}, and Wan2.1~\cite{wang2025wan} demonstrate strong spatiotemporal coherence and high video quality. These methods have been widely applied in fields including animation generation~\cite{he2023animate, hu2024animate}, video editing~\cite{zhang2025instructvedit, wang2024taming}, and world modeling~\cite{meng2024towards,he2025pre}.

\begin{figure*}[!t]
    \centering
    \includegraphics[width=0.95 \linewidth]{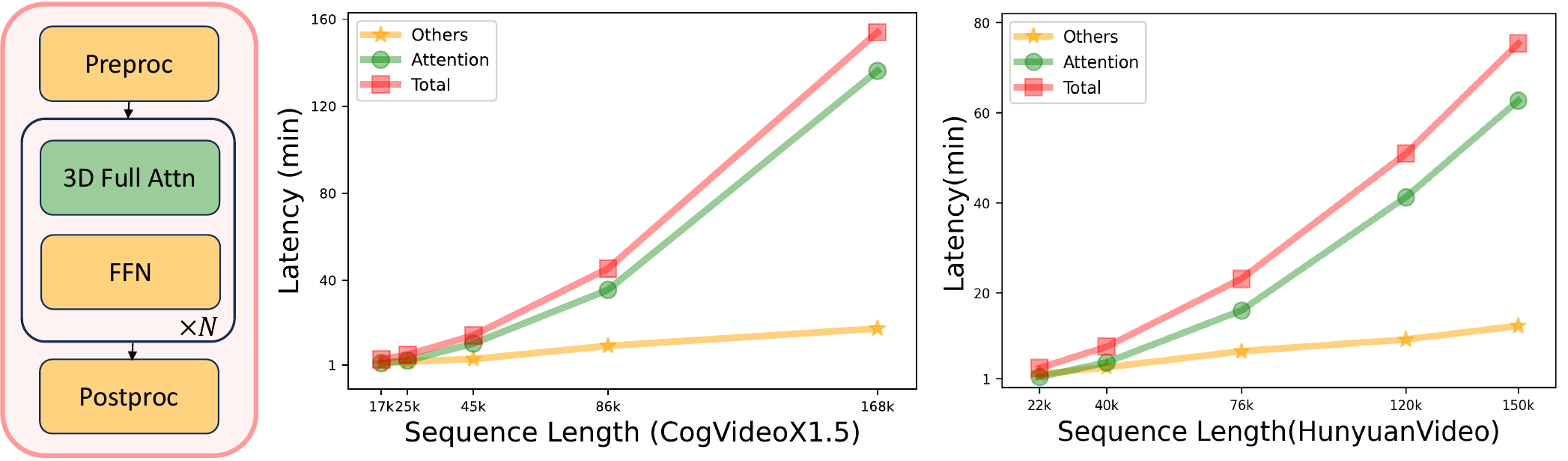}
    \caption{The architecture of vDiT and inference latency analysis of its two variants, CogVideoX1.5 and HunyuanVideo, across different components. The latency of the attention module dominates under long sequence settings, and its proportion increases as the sequence length grows.}
    \label{fig:attentionratio}
\end{figure*}

Although 3D full-attention vDiT models demonstrate strong video generation performance and are widely adopted, they suffer from high computational costs and large inference latency. For instance, generating a 5-second 720p video at 24 fps using the HunyuanVideo model on a single NVIDIA A800 GPU takes approximately fifty minutes. This inefficiency primarily results from the joint spatiotemporal tokenization process, which generates up to 120k tokens in this setting. Given that attention complexity scales quadratically with sequence length~\cite{attention2017vaswani}, this leads to a substantial computational burden. As shown in Figure~\ref{fig:attentionratio}, in the classical model CogVideoX1.5, which is based on the vDiT architecture, attention accounts for 77\% of the total latency at 86k tokens. Specifically, for HunyuanVideo with 120k tokens, attention accounts for 81\% of the total inference latency, and this proportion increases with longer sequence length. Thus, 3D full attention is the primary bottleneck in inference efficiency for vDiT-based video generation.

Fortunately, the 3D full attention mechanism exhibits significant redundancy despite its considerable computational cost. First, we observe that some attention heads in vDiT are redundant, as skipping their computations has minimal effect on the final output. Second, redundancy is also present in the computation of the attention map, namely in the $\bm{QK}^T$ product. We find that vDiT attention maps commonly follow four distinct patterns: full attention, diagonal sparsity, multi-diagonal sparsity, and vertical-strip sparsity. The latter three patterns suggest that computing the full attention map is often unnecessary. Further experiments reveal that these sparse patterns remain stable across different input texts and are primarily determined by the position of attention within the vDiT architecture. This fixed redundancy provides a strong basis for optimization.

Building on these findings, we propose Sparse-vDiT, a sparse method designed to accelerate vDiT for video generation. To reduce redundancy among attention heads, we introduce a head skipping strategy. We observe that vDiT’s attention maps commonly follow three sparse patterns: diagonal, multi-diagonal, and vertical-stripes. To enable actual speedup, we design predefined kernels tailored to each pattern. Since these sparsity patterns are relatively fixed and input-invariant, we develop an offline sparse diffusion search algorithm that identifies the optimal attention pattern for each head using only a small number of samples. After the search, the computation pattern of each head is fixed. We then group heads with the same sparsity pattern within each layer and fuse them to further accelerate inference by leveraging their fixed structure. We conducted experiments on three widely used vDiT-based models: CogVideoX1.5, HunyuanVideo, and Wan2.1. On CogVideoX1.5, Sparse-vDiT achieved a 2.09$\times$ reduction in theoretical FLOPs and a 1.76$\times$ end-to-end speedup in real, while keeping the LPIPS score low at 0.14, and even outperforming the baseline in the ImageQual metric. On HunyuanVideo, Sparse-vDiT achieved a 2.38$\times$ reduction in theoretical FLOPs and a 1.85$\times$ speedup, with generation quality reaching SSIM = 0.87 and PSNR = 27.03. On Wan2.1, Sparse-vDiT achieved a 1.67$\times$ reduction in theoretical FLOPs and a 1.58$\times$ speedup, with generation quality reaching SSIM = 0.80 and PSNR = 22.59. These results indicate that Sparse-vDiT effectively balances computational efficiency and generation quality.

The contributions of our paper are as follows:

\begin{itemize}[leftmargin=1em]
\item We find that attention heads in vDiT are partly redundant. Meanwhile, many heads often exhibit recurring sparse attention patterns, including diagonal sparsity, multi-diagonal sparsity, and vertical-stripe sparsity. These patterns are consistent across different inputs but are closely related to the attention position within the vDiT architecture.
\item Building on these insights, we propose Sparse-vDiT, which accelerates vDiT by skipping redundant heads and applying pattern-aligned sparse attention kernels. It introduces an offline sparse diffusion search that selects the optimal sparse mode for each head using a small number of samples, followed by intra-layer fusion of heads with identical attention patterns to enhance inference efficiency.
\item Sparse-vDiT achieves 2.09$\times$, 2.38$\times$ and 1.67$\times$ theoretical FLOPs reduction on CogVideoX1.5, HunyuanVideo and Wan2.1, respectively. It also delivers 1.76$\times$, 1.85$\times$, and 1.58$\times$ end-to-end video generation speedups while maintaining comparable generation quality, with PSNR scores of 24.13, 27.09, and 22.59. Sparse-vDiT consistently outperforms existing state-of-the-art (SOTA) methods, such as SVG and MInference.
\end{itemize} 

\section{Related Work}
\textbf{Efficient Diffusion Model.}\
Diffusion models are inherently slow because of their iterative denoising process, leading to growing interest in accelerating inference. 
Existing approaches include pruning methods~\cite{diff-pruning2023fang, pdpruner2024castells} that reduce model parameters, quantization techniques~\cite{ptq4dit2024wu, ptq4dm2023shang, zhao2024fp4} that decrease parameter bit-width and computational overhead, and caching strategies~\cite{ma2024deepcache, chen2024delta, mddit2024shen} that trade memory for computation speed. However, most of these methods are primarily designed for image generation, with relatively few acceleration methods specifically tailored for video diffusion models.
For video diffusion, techniques like PAB~\cite{pab2024zhao}, TeaCache~\cite{teacache2024liu}, FasterCache~\cite{fastercache2024lv}, and AdaCache~\cite{adacache2024kahatapitiya} reuse features by exploiting the similarity between adjacent denoising steps. Other methods reduce the number of timesteps using distillation~\cite{onediffusion2025lin,zhai2024motion} or compress latent spaces using high-ratio VAEs~\cite{reducio2024tian}. In contrast, our approach accelerates inference by exploiting the sparsity in vDiT’s attention.

\textbf{Efficient Attention Mechanism.}\
The attention~\cite{attention2017vaswani} is central to transformers but suffers from quadratic complexity in sequence length, limiting efficiency in long sequences. To address this, various sparse attention methods have been proposed. In traditional vision, Swin Transformer~\cite{swin2021liu}, NAT~\cite{nat2023hassani}, and Sparse Transformers~\cite{sparse2019child} restrict attention to local windows. Similarly, Longformer~\cite{longformer2020beltagy} applies windowed attention in NLP. Large language models~\cite{llama2023touvron} have identified attention sink phenomena~\cite{streaming2023xiao, duoattention2024xiao}, introducing streaming attention that combines sink masking with windowing. Later works, such as MInference~\cite{minference2024jiang} and FlexPrefill~\cite{flexprefill2025lai}, explore diverse static and dynamic sparse patterns. In diffusion models, DiTFastAttn~\cite{ditfastattn2024yuan, ditfastattnv22025zhang} noted strong local neighbor attention in DiTs, enabling acceleration via windowed attention and cached contexts. CLEAR~\cite{clear2024liu}, DiG~\cite{dig2024zhu}, and SANA~\cite{sana2024xie} further exploit the sparsity of the attention mechanism to achieve linearized computation. For video diffusion, Efficient-vDiT \cite{efficientvdit2025ding} observed that each frame in the attention primarily attends to a fixed set of other frames. This observation introduces tile-based attention to linearized computation. SVG \cite{svg2025xi} identified spatiotemporal sparsity in video attention and optimized attention computation through data reordering and an online scheme.
However, this paper thoroughly reveals multiple patterns and invariances of redundancy in vDiT attention. Based on these findings, we propose an offline sparse acceleration framework that integrates head skipping with three attention sparsity patterns. Considering the fixed nature of offline optimization, fusion optimization is performed on a fixed attention pattern at each attention layer.

\section{Preliminary}

\textbf{Full Attention.}
The multi-head attention mechanism~\cite{attention2017vaswani} constitutes a fundamental building block in vDiT. Let the input hidden features be denoted as $\bm{I}\in \mathbb{R}^{B\times N\times D}$, where $B$ is the batch size, $N$ the number of tokens, and $D$ the original feature dimension. Through learnable linear projections, $\bm{I}$ is transformed into three tensors: query ($\bm{Q}$), key ($\bm{K}$) and value ($\bm{V}$). Each of these tensors has dimensions $\mathbb{R}^{B\times H\times N\times d}$, where $H$ denotes the number of attention heads, and $d=D/H$ represents the reduced feature dimension per head. The attention outputs refined features $\bm{O}\in \mathbb{R}^{B\times N\times D}$ preserving the original dimension of $\bm{I}$. The attention transformation process is defined as follows: for each head $h\in \{1,...,H\}$,
\begin{equation}
    Attention(\bm{Q}_h,\bm{K}_h,\bm{V}_h)=softmax(\bm{Q}_h\bm{K}_h^T / \sqrt{d})\bm{V}_h\in \mathbb{R}^{B\times N\times d},
    \label{eq:attention}
\end{equation}
where $\bm{Q}_h, \bm{K}_h, \bm{V}_h$ are slice operations on the head dimension. Then, merging along the head dimension yields the final output $\bm{O}$ of the attention. For the full attention mechanism, the entire process described above is executed.

\textbf{Sparse Attention.}
In Eq~\ref{eq:attention}, $softmax(\bm{Q}_h\bm{K}_h^T/\sqrt{d})$ is known as the attention map, where each value represents how much one token attends to another at the corresponding position. Since its computational complexity is $\mathcal{O}(N^2)$, generating the attention map takes up most of the computation in the attention mechanism. However, in practice, a token usually attends to only a small number of other tokens, rather than maintaining global attention. This results in most values in the attention map being close to zero, showing strong sparsity. In most cases, it is sufficient to compute only the dense regions of the attention map to obtain a sufficiently accurate result. If the sparsity pattern of the attention map is structured, computations involving sparse regions can be omitted at the hardware level using Triton~\cite{tillet2019triton} or CUDA, enabling practical acceleration.


\section{Method}
\begin{wraptable}{r}{0.5\textwidth}
\small
\centering
\vspace{-50pt}
\caption{Quantitative impact of skipping different ratios of attention heads on the final generation.}
\resizebox{\linewidth}{!}{
\begin{tabular}{c|ccc}
\toprule
\textbf{CogVideoX1.5}& \textbf{PSNR $\uparrow$}   & \textbf{SSIM $\uparrow$}   & \textbf{LPIPS $\downarrow$}  \\
\midrule
skipping 1\%                   & 36.62& 0.96& 0.01\\
skipping 3\%                   & 33.31& 0.95& 0.02\\
skipping 6\%                   & 30.02& 0.92& 0.04\\
skipping 10\%                  & 26.87& 0.85& 0.09\\
\midrule
\textbf{HunyuanVideo} & \textbf{PSNR $\uparrow$}   & \textbf{SSIM $\uparrow$}   & \textbf{LPIPS $\downarrow$}  \\
\midrule
skipping 1\%                   & 31.84& 0.95& 0.02\\
skipping 3\%                   & 28.94& 0.91& 0.06\\
skipping 6\%                   & 24.21& 0.81& 0.12\\
skipping 10\%                  & 17.98& 0.72& 0.22\\
\bottomrule
\end{tabular}}
\label{tab:skip}
\vspace{-30pt}
\end{wraptable}

\subsection{Attention Mechanism in vDiT}
In the following, we present the attention mechanism employed in vDiT. We first describe the distinctive layout of attention maps tailored for video generation. Next, we demonstrate that the attention mechanism exhibits substantial redundancy. Finally, we show that this redundancy is largely intrinsic to the model architecture and remains relatively insensitive to variations in the input.

\subsubsection{Attention Map in vDiT.}
\begin{figure*}[t]
    \centering
    \includegraphics[width=0.99 \linewidth]{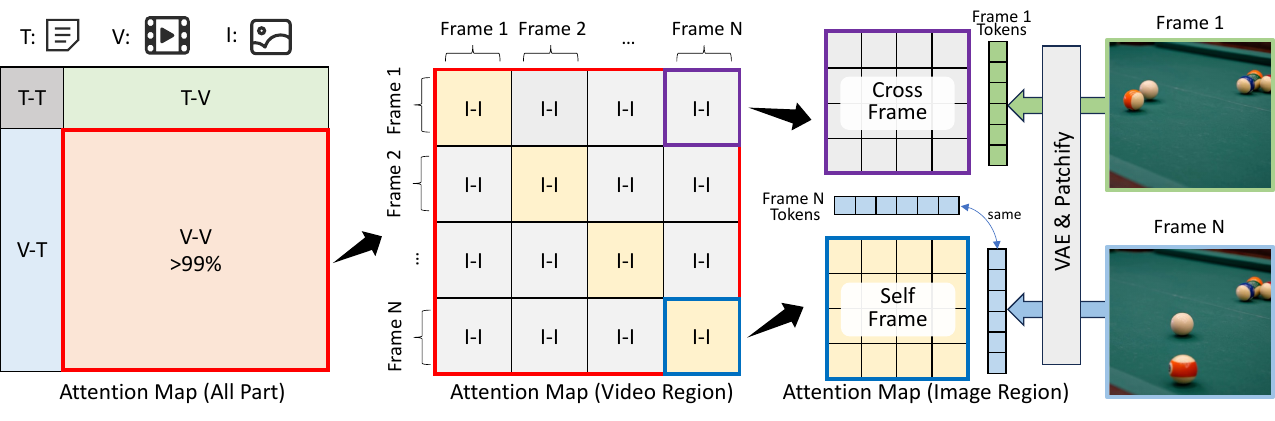}
    \caption{Visualization of the vDiT attention map showing four interaction regions. The dominant V-V region has diagonal blocks for self-frame and off-diagonal blocks for cross-frame interactions.}
    \vspace{-10pt}
    \label{fig:attentionmap}
\end{figure*}

Current mainstream vDiT models, such as CogVideoX and HunyuanVideo, mainly adopt the MM-DiT paradigm~\cite{esser2024sd3}. In this design, the token sequence is formed by concatenating text tokens and video tokens, and the corresponding attention map is shown on the left side of Figure~\ref{fig:attentionmap}. The attention map is divided into four parts based on token type and position: T–T, T–V, V–T, and V–V, where T denotes text tokens and V denotes video tokens. Text tokens make up only a small portion of the sequence, while video tokens account for over 99\%. In the V–V region (the middle part of Figure~\ref{fig:attentionmap}), video tokens are arranged in the temporal order of frames. As a result, the diagonal blocks correspond to self-frame interactions among image (frame) tokens. In contrast, the off-diagonal blocks correspond to cross-frame interactions, as illustrated on the right part of Figure~\ref{fig:attentionmap}.

\subsubsection{Analyzing Attention Redundancy in vDiT}
We find that attention in vDiT contains considerable redundancy. Some attention heads are non-essential, and skipping them results in minimal performance loss. Moreover, the attention maps exhibit patterns of structured sparsity, which can be exploited to enable efficient sparse computation.

\textbf{Head Skipping.} Not all attention heads in vDiT contribute equally to performance. Based on a minimum mean squared error (MSE) criterion, we evaluate head skipping on CogVideoX1.5 and HunyuanVideo. As shown in Table~\ref{tab:skip}, in CogVideoX1.5, skipping 6\% of the attention heads preserves generation quality comparable to the original model. In HunyuanVideo, skipping 3\% of the heads similarly causes little degradation in video quality. These results indicate that certain attention heads in vDiT are redundant, suggesting that head skipping may be a practical means to improve efficiency. However, relying solely on skipping is insufficient to achieve high efficiency. As a coarse-grained method, it results in performance degradation beyond a certain threshold. As shown in Table~\ref{tab:skip}, both models exhibit noticeable degradation when the skip ratio reaches 10\%. Therefore, a more fine-grained strategy is required to achieve a greater speedup.

\begin{figure*}[t]
    \centering
    \includegraphics[width=0.99 \linewidth]{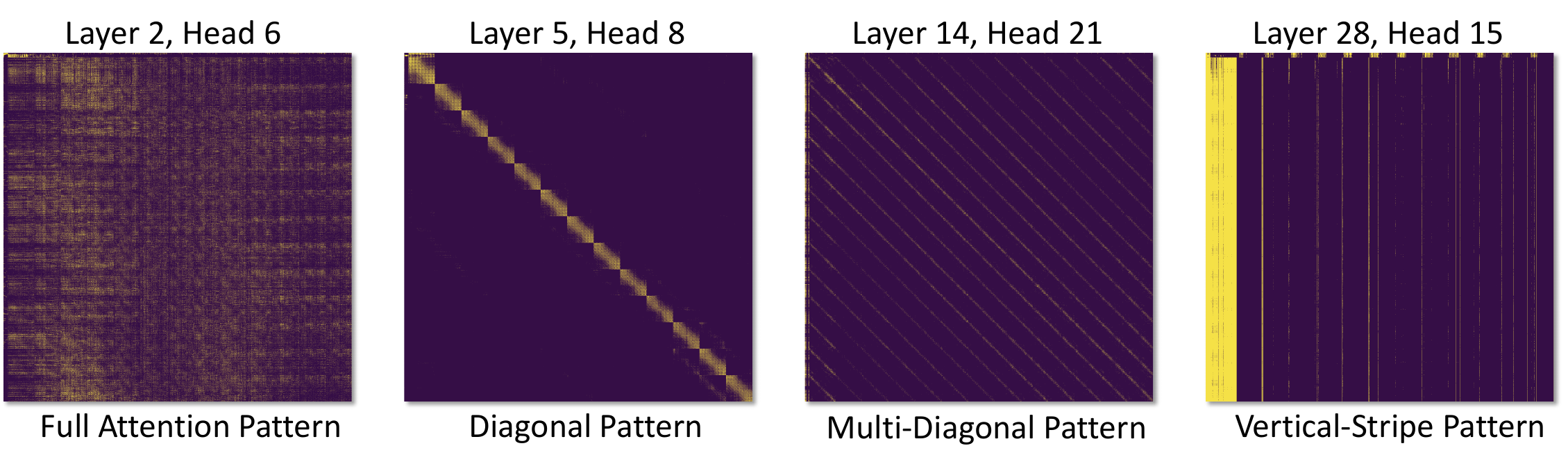}
    \caption{Visualization of the four recurring attention patterns in vDiT.}
    \label{fig:attentionpattern}
    \vspace{-5pt}
\end{figure*}

Given that the sparsity of attention maps can improve the efficiency of transformer models, we conduct an in-depth analysis of the attention map in vDiT. Taking CogVideoX as an example, we visualize its attention maps in Figure~\ref{fig:attentionpattern} and identify four recurring patterns:

\textbf{Full Attention Pattern.} The attention values are evenly distributed, indicating global interactions among tokens. Applying sparse computation to such dense patterns often degrades performance, making efficiency optimization difficult.

\textbf{Diagonal Pattern.} Large values appear along the main diagonal, representing interactions among neighboring tokens within the same frame (as shown in Figure~\ref{fig:attentionmap}). This pattern reflects the model’s ability to capture self-frame structure. Since most off-diagonal values are close to zero, the full attention can be well approximated by computing only the diagonal elements of the attention map. This structured sparsity allows for efficient acceleration using window attention~\cite{longformer2020beltagy}.

\textbf{Multi-Diagonal Pattern.} Large values are distributed along multiple evenly spaced diagonals. These diagonals align with the diagonal blocks in the I-I region of Figure~\ref{fig:attentionmap}, indicating strong attention between tokens at nearby spatial positions across different frames. Therefore, this pattern is associated with vDiT’s ability to model cross-frame consistency. By rearranging tokens~\cite{svg2025xi}, this pattern can be transformed into a diagonal structure suitable for optimization with window attention.

\textbf{Vertical-Stripe Pattern.} In the attention map, large values form a vertical stripe pattern, suggesting the presence of global tokens that strongly attend to all others in vDiT. This structured sparsity also enables efficient computation by a sparse kernel.

\begin{figure*}[!h]
    \centering
    \includegraphics[width=0.99 \linewidth]{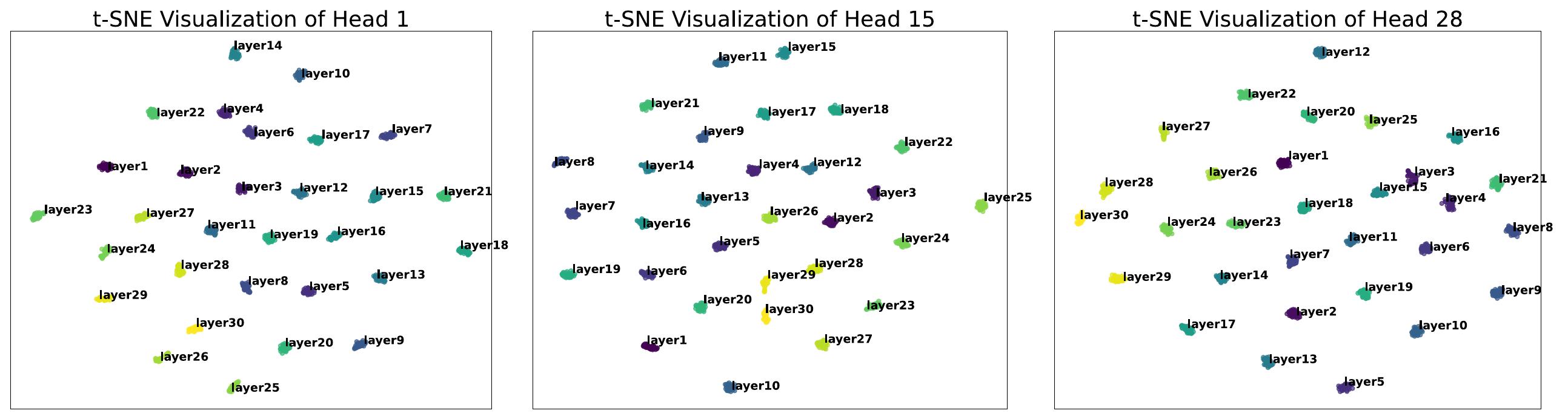}
    \caption{t-SNE visualization of attention patterns along the head dimension on a VBench subset, with different layers indicated by distinct colors. Patterns from different prompts exhibit clustering.}
    \label{fig:tsnepattern}
\end{figure*}
\subsubsection{Invariant Property of Attention Patterns}
We revealed the presence of diverse attention patterns in vDiT above. We further observe that these patterns are strongly correlated with the depth of the attention layers, while being largely independent of the input text. To verify this, we randomly sampled 50 diverse prompts from VBench as a subset and used them to generate videos. For each layer and each attention head in vDiT, we saved the corresponding attention maps. Since we only needed to determine the pattern types, we stored the maps as memory-efficient image files. We then used a ResNet50 to extract high-dimensional features from the images and applied t-SNE to project them into a 2D space along the head dimension. The results are shown in Figure~\ref{fig:tsnepattern}, where different colors represent different layers. We observed that, regardless of the head, the attention patterns from different layers form distinct clusters, while those from different prompts tend to cluster together. This confirms that the attention patterns exhibit strong correlations with attention position in vDiT but are minimally affected by the input content.

\begin{figure*}[t]
    \centering
    \includegraphics[width=0.99 \linewidth]{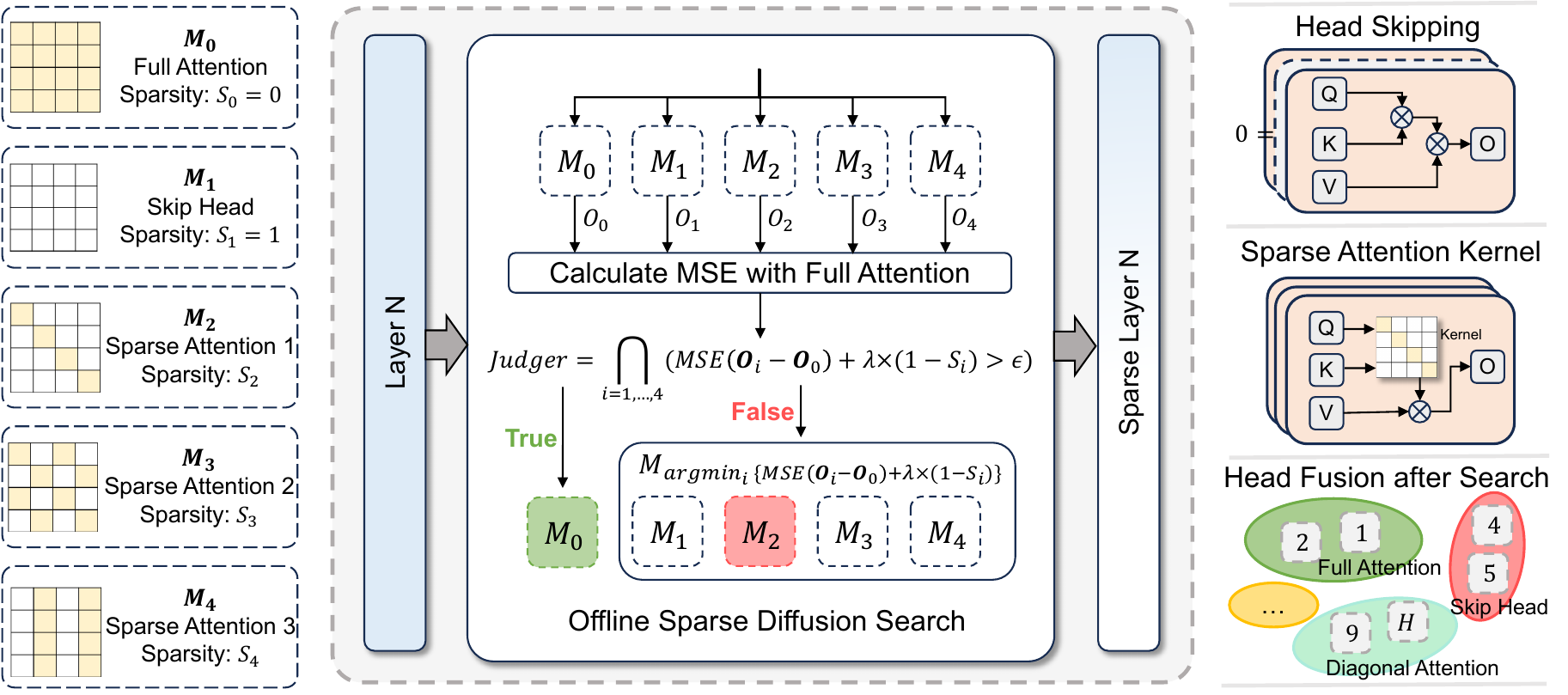}
    \caption{\textbf{Overview of the Sparse-vDiT.} We first predefine five types of attention mode $M_{0:4}$. Then, using an offline sparse diffusion search algorithm, we select the best attention mode for each layer and head in vDiT. After the search, for heads set to skip attention, we set their outputs to zero. For the three sparse attention patterns, we create specialized sparse attention kernels to speed up computation. Finally, heads within the same layer that use the same attention mode are fused to improve efficiency.}
    \label{fig:main}
    \vspace{-10pt}
\end{figure*}

\subsection{Sparse-vDiT: A Sparse Acceleration Framework for vDiT}
In the previous part, we identified two types of redundancy in the attention mechanism of vDiT: redundancy within the attention heads and redundancy in the attention map computation. We also found that this redundancy is intrinsic to vDiT and only weakly dependent on the input text. Based on these findings, we introduce Sparse-vDiT, a sparse acceleration method designed for vDiT. This method determines the most effective sparse strategy for each head in each layer through offline search, resulting in acceleration. The overall structure of Sparse-vDiT is illustrated in Figure~\ref{fig:main}.

\textbf{Sparse Computation Pre-definition.} To reduce redundancy in the attention head, we apply the skip strategy $M_1$, which bypasses the entire process in Eq~\ref{eq:attention}. To maintain consistent output dimensions, the attention output is set to zero. The sparsity of the skip strategy is defined as $S_1 = 1$, while the sparsity of full attention $M_0$ is $S_0 = 0$. Regarding the three sparse forms $M_i (i=2,3,4)$ shown in Figure~\ref{fig:attentionpattern}, we have designed specific sparse kernels to reduce the computation of $softmax(\bm{QK}^T/\sqrt{d})$. The sparsity of these kernels is determined by the ratio of actual computation blocks to the total number of blocks, denoted as $S_i (i=2,3,4)$, as shown in Figure~\ref{fig:main}. In the Sparse-vDiT framework, the sparsity of these kernels is predefined and treated as a fixed constant.

\textbf{Offline Sparse Diffusion Search.} In vDiT, different heads at various layers exhibit distinct attention patterns. Given the set $\bm{M} = \{ M_i (i=0,...,4)\}$ of attention computation modes, the challenge lies in selecting the most appropriate mode for each head. In Sparse-vDiT, we propose an offline sparse diffusion search method to address this. As shown in Figure~\ref{fig:main}, for each layer in every step of vDiT, we pass the inputs through $M_0$ to $M_4$, obtaining the corresponding hidden state results $\bm{O}_0$ to $\bm{O}_4$. We then compute the MSE distances between $\bm{O}_1$ to $\bm{O}_4$ and $\bm{O}_0$, that is, $MSE(\bm{O}_i-\bm{O}_0),i=1,...,4$, which represent the loss introduced by the sparse attention computation. Our final loss is
\begin{equation}
L_i=MSE(\bm{O}_i-\bm{O}_0) + \lambda \times (1-S_i), 
\end{equation}
where the sparsity penalty is added and $\lambda$ balances quality and computational cost. If all losses in $\bm{L}=\{L_i (i=1,...,4)\}$ exceed the desired threshold $\epsilon$, the head retains full attention. Otherwise, the sparse mode with the smallest loss replaces full attention. The specific formulation is as follows:

\begin{equation}
Attention(\bm{Q},\bm{K},\bm{V},\bm{M})=
\begin{cases}
M_0(\bm{Q},\bm{K},\bm{V})& \text{, if}\space \bigcap\limits_{i=1,...,4}(L_i>\epsilon)\\
M_{argmin_i\{L_i\}}(\bm{Q},\bm{K},\bm{V})& \text{, otherwise}
\end{cases}
\end{equation}
where $\epsilon$ controls the overall sparsity ratio during inference. As discussed in the previous part, vDiT's sparse attention pattern is inherent after pretraining and largely independent of input types. Thus, the search in Sparse-vDiT is offline and requires only a small number of input samples. Once the search is completed, the sparse modes for the entire inference process are fixed. This fixity allows heads with the same sparse mode within a layer to be fused, further accelerating the inference.

\vspace{-10pt}
\section{Experiment}

\subsection{Experimental Settings}
\textbf{Pretrained Model.}
To evaluate the effectiveness of Sparse-vDiT, we conducted text-to-video generation experiments using three leading open-source pretrained vDiT models: CogVideoX1.5~\cite{cogvideox2024yang}, HunyuanVideo~\cite{kong2024hunyuanvideo}, and Wan2.1~\cite{wang2025wan}. CogVideoX1.5 generates 81 frames at a resolution of 1360$\times$768, while HunyuanVideo generates 129 frames at 1280$\times$720. In the latent space encoded by the 3D-VAE, the vDiT in CogVideoX1.5 processes 45,106 tokens, including 226 text tokens and 11 video frames with 4,080 tokens each. The vDiT in HunyuanVideo processes 119,056 tokens, including 256 text tokens and 33 frames with 3,600 tokens each. And the vDiT in Wan2.1 processes 75,600 tokens, including 21 frames with 3,600 tokens each.

\textbf{Dataset \& Evaluation Metrics.}
We adopted a comprehensive evaluation framework covering both video generation quality and efficiency. For quality evaluation, we used three types of metrics. The first category measures reconstruction fidelity after inference acceleration, including Peak Signal-to-Noise Ratio (PSNR)~\cite{pab2024zhao}, Structural Similarity Index Measure (SSIM)~\cite{wang2002ssim}, and Learned Perceptual Image Patch Similarity (LPIPS)~\cite{zhang2018LPIPS}. The remaining two categories assess frame-level visual quality and temporal consistency, using the Imaging Quality (ImageQual) and Subject Consistency (SubConsist) metrics from the VBench~\cite{huang2024vbench}. For efficiency evaluation, we considered theoretical FLOPS, actual inference latency, and the speedup relative to the pretrain model. Regarding evaluation datasets, we followed the original protocol of CogVideoX~\cite{cogvideox2024yang}, using prompts from the GPT-enhanced version of VBench. For HunyuanVideo, we used prompts from the Penguin Video Benchmark~\cite{kong2024hunyuanvideo}.

\textbf{Baseline.}
We compared several existing acceleration methods for vDiT, including both classical approaches and state-of-the-art techniques. These methods include MInference~\cite{minference2024jiang}, a classical sparse acceleration technique migrated from large language models. WinAttn~\cite{longformer2020beltagy}, which applies sparse acceleration along both temporal and spatial dimensions of video. SVG~\cite{svg2025xi}, the current state-of-the-art method for sparse accelerating vDiTs, and PAB~\cite{pab2024zhao}, a caching-based method designed specifically for video diffusion models.

\textbf{Implementation Details.}
The baselines MInference, PAB, and SVG are implemented using their official code and configurations. Since PAB only provides code for CogVideo, we do not include it in the evaluation on HunyuanVideo. The window sizes for WinAttn-Spatial and WinAttn-Temporal follow the settings used in SVG. In SVG, full attention is applied during the first 10 steps, and we follow the same setup for all baselines. However, this constraint is not required for Sparse-vDiT on CogVideoX1.5. SVG also applies full attention to the first two layers of vDiT, and we adopt the same configuration for our baselines, although it is unnecessary for Sparse-vDiT. Both CogVideoX1.5 and HunyuanVideo inference results were obtained on a single NVIDIA A800 GPU, while Wan2.1 was obtained on a single NVIDIA H800 with a batch size of 1.

\begin{table*}[!t]
    \centering
    \caption{Comparison of video generation quality and efficiency between Sparse-vDiT and the baseline. XBench refers to VBench for CogVideoX1.5 and Wan2.1 evaluation and Penguin Video Bench for HunyuanVideo. CogVideoX1.5 \& HunyuanVideo on single A800, Wan2.1 on H800, batch size 1.}
    \resizebox{\linewidth}{!}{
    \begin{tabular}{l|cccccccc}
    \toprule
    
    \multirow{3}{*}{\textbf{Method}} & \multicolumn{3}{c}{\textbf{Against Original}}   & \multicolumn{2}{c}{\textbf{XBench Score}} & \multirow{3}{*}{\textbf{PFLOPS} ($\downarrow$)}& \multirow{3}{*}{\textbf{Latency} ($\downarrow$)} & \multirow{3}{*}{\textbf{Speedup} ($\uparrow$)} \\
     \cmidrule(lr){2-4} \cmidrule(lr){5-6}
     & \textbf{SSIM} ($\uparrow$) & \textbf{PSNR} ($\uparrow$) & \textbf{LPIPS} ($\downarrow$) & \textbf{ImageQual } ($\uparrow$) & \textbf{SubConsist} ($\uparrow$) &    &   &      \\
    \midrule
    \textbf{CogVideoX1.5}~\cite{cogvideox2024yang}& -   & -  & -   & 63.28\%   & 92.96\% & 147.87& 901s   & 1.00$\times$\\
    
    MInference~\cite{minference2024jiang}  & 0.61& 14.63& 0.37& 56.04\%  & 87.12\%   & 84.89& 634s & 1.42$\times$   \\
    WinAttn (Spatial)~\cite{longformer2020beltagy}& 0.64& 19.07& 0.32& \textbf{64.84\%}& 90.92\%   & 72.34& 531s& 1.69$\times$\\
    WinAttn (Temporal)~\cite{longformer2020beltagy}& 0.69& 19.64& 0.28& 63.69\%   & 92.66\%   & 72.34& 537s& 1.67$\times$\\
    PAB~\cite{pab2024zhao}  & 0.72& 20.93& 0.23& 59.03\%   & 92.38\%   & 105.88& 630s   & 1.43$\times$ \\
    SVG~\cite{svg2025xi}& 0.75& 21.92& 0.22& 63.11\%   & 92.49\%   & 74.57& 550s& 1.64$\times$\\
     \rowcolor[HTML]{EFEFEF} 
    Sparse-vDiT (Ours)& \textbf{0.82}& \textbf{24.13}& \textbf{0.14}& 63.45\%& \textbf{92.66\%}&  \textbf{70.69}& \textbf{511s}& \textbf{1.76$\times$}\\
    
    \midrule
    \textbf{HunyuanVideo}~\cite{kong2024hunyuanvideo}& -   & -  & -   & 67.28\%   & 96.79\%   & 612.37& 3166s   & 1.00$\times$\\
    MInference~\cite{minference2024jiang}  & 0.64& 19.23& 0.43& 60.53\%  & 88.96\%   & 293.87& 2042s & 1.55$\times$   \\
    WinAttn (Spatial)~\cite{longformer2020beltagy}& 0.56& 17.81& 0.56& 63.55\% & 90.26\%   & 258.84& 1755s& 1.80$\times$\\
    WinAttn (Temporal)~\cite{longformer2020beltagy}& 0.80& 23.76& 0.22& \textbf{67.32\%}& 96.38\%   & 258.84& 1764s& 1.79$\times$\\
    SVG~\cite{svg2025xi}& 0.86& 26.83& 0.14& 67.06\%   & 96.54\%   & 259.79& 1802s& 1.75$\times$\\
     \rowcolor[HTML]{EFEFEF} 
    Sparse-vDiT (Ours)& \textbf{0.87}& \textbf{27.09}& \textbf{0.12}& 67.13\%& \textbf{96.69\%}&  \textbf{257.09}& \textbf{1715s}& \textbf{1.85$\times$}\\

    \midrule
    \textbf{Wan2.1}~\cite{wang2025wan}& -   & -  & -   & 67.61\%& 91.95\%& 660.49& 1935s& 1.00$\times$\\
    MInference~\cite{minference2024jiang}  & 0.62& 15.49& 0.36& 63.29\%& 89.32\%& 469.79& 1453s& $1.33\times$\\
    WinAttn (Spatial)~\cite{longformer2020beltagy}& 0.68& 19.14& 0.25& 67.27\%& 91.34\%& 401.21& 1265s& $1.53\times$\\
    WinAttn (Temporal)~\cite{longformer2020beltagy}& 0.73& 20.29& 0.21& \textbf{67.40\%}& \textbf{91.47\%}& 401.21& 1280s& $1.51\times$\\
    SVG~\cite{svg2025xi}& 0.78& 21.96& 0.18& 67.18\%& 91.27\%& 403.50& 1298s& $1.49\times$\\
     \rowcolor[HTML]{EFEFEF} 
    Sparse-vDiT (Ours)& \textbf{0.80}& \textbf{22.59}& \textbf{0.16}& 67.35\%& 91.39\%&  \textbf{397.39}& \textbf{1228s}& \textbf{1.58$\times$}\\
    
    \bottomrule
    \end{tabular}
    }
    \vspace{-10pt}
    \label{tab:main_results}
\end{table*}

\subsection{Experimental Results Analysis}
\begin{figure*}[t]
    \centering
    \includegraphics[width=0.99 \linewidth]{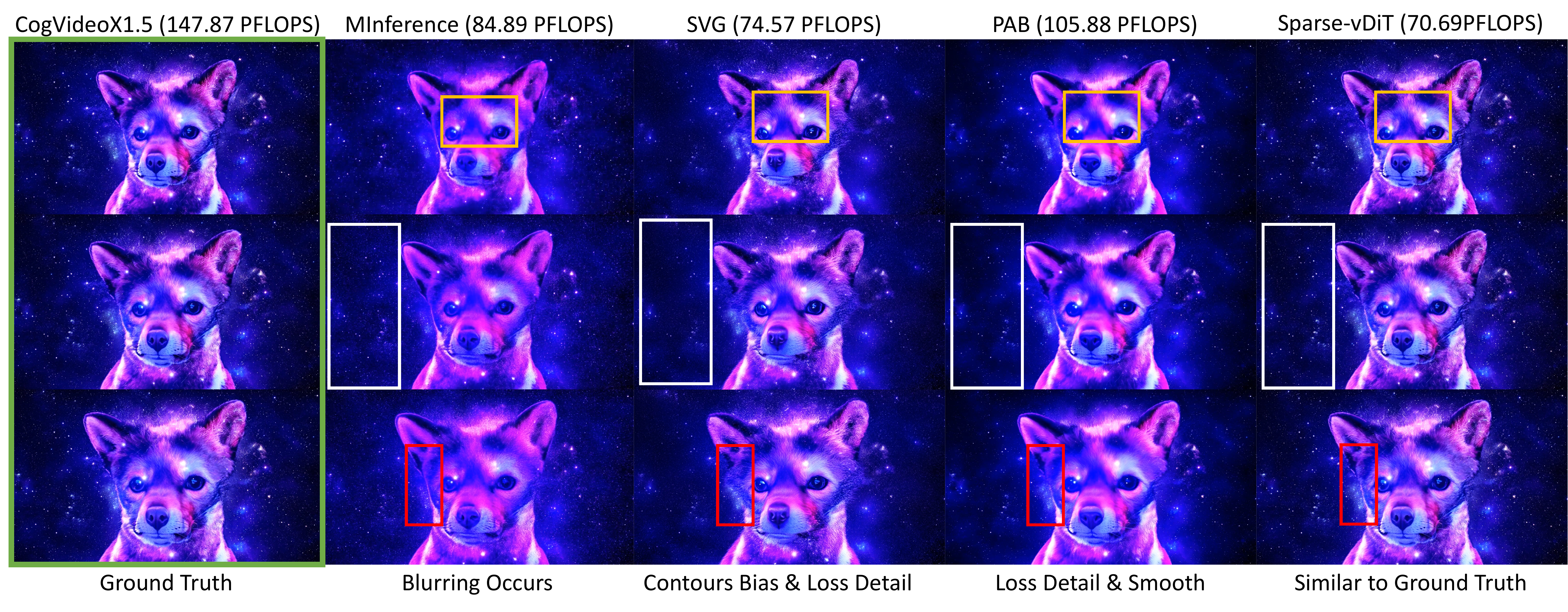}
    \caption{Visual comparison between the proposed Sparse-vDiT and the baseline method. The green box indicates the ground truth. Yellow boxes highlight differences in blurriness and smoothness. White boxes highlight differences in fine details, while red boxes emphasize contour comparisons.}
    \label{fig:compare}
    \vspace{-15pt}
\end{figure*}

The qualitative and quantitative results are shown in Figure~\ref{fig:compare} and Table~\ref{tab:main_results}, respectively. Both consistently demonstrate that Sparse-vDiT effectively accelerates video diffusion models without compromising the quality of generation. This can be explained as follows.

\textbf{Reconstruction Fidelity.} On both CogVideoX1.5 and HunyuanVideo, Sparse-vDiT achieves the best performance across all fidelity metrics. For CogVideoX1.5, Sparse-vDiT yields an SSIM of 0.82, significantly higher than the closest baseline, SVG (0.75), and substantially higher than earlier sparse methods, such as MInference (0.61) and PAB (0.72). Similarly, the PSNR for Sparse-vDiT is 24.13 dB, surpassing all baselines, with the suboptimal result from SVG at 21.92 dB. Most notably, Sparse-vDiT achieves a substantially lower LPIPS score (0.14), indicating greater perceptual similarity to the original outputs. The trends hold consistently on HunyuanVideo, where Sparse-vDiT again records the highest SSIM (0.87) and PSNR (27.09), along with the lowest LPIPS (0.12). The margins are particularly significant compared to early techniques such as WinAttn (Temporal), which, while effective (SSIM: 0.76, LPIPS: 0.22), still underperforms relative to Sparse-vDiT. These results confirm the strong preservation of spatial and perceptual detail after applying our acceleration scheme.

\textbf{Visual Quality.}
The ImageQual score from the VBench benchmark quantifies the frame-level visual quality as judged by pretrained evaluation models. Sparse-vDiT performs on par with or better than most baselines, achieving 63.45\% on CogVideoX1.5 and 67.13\% on HunyuanVideo. Although WinAttn (Spatial) slightly surpasses Sparse-vDiT in ImageQual on CogVideoX1.5 (64.84\%), it comes with lower fidelity scores and higher LPIPS, suggesting a potential overfitting to local texture patterns at the cost of content preservation. On HunyuanVideo, Sparse-vDiT delivers ImageQual scores highly comparable to the best-performing methods, including SVG (67.06\%) and WinAttn (Temporal) (67.32\%). These results indicate that Sparse-vDiT maintains competitive frame-level realism while significantly outperforming others in reconstructive metrics, highlighting its balanced and robust generation performance.

\textbf{Temporal Consistency.}
Temporal coherence is critical in video generation, and the SubConsist metric evaluates the consistency of subjects and motion across frames. Sparse-vDiT delivers state-of-the-art temporal stability in both benchmarks. On CogVideoX1.5, its SubConsist score reaches 92.66\%, on par with the strongest existing methods, including WinAttn (Temporal) and PAB. On HunyuanVideo, Sparse-vDiT attains 96.69\%, closely matching the best score of 96.79\% from the original unaccelerated model. This observation is particularly important because many acceleration methods compromise temporal stability in favor of spatial quality. The ability of Sparse-vDiT to achieve high consistency while also delivering best-in-class fidelity underscores the effectiveness of its sparse acceleration strategy. By preserving computation in more temporally sensitive heads, Sparse-vDiT minimizes temporal artifacts common in other sparsity approaches.

\textbf{Visualization.} Figure~\ref{fig:compare} shows a visual comparison between the video results generated by Sparse-vDiT and those from the top three baseline methods. We observe that MInference produces blurry results, while PAB shows over-smoothing, as indicated by the yellow box in the first row. Both SVG and PAB lose some fine details, as shown in the white box in the second row. For object contours, SVG exhibits a slight misalignment, as indicated by the red box in the third row. In contrast, our method remains closely aligned with the pretrained model in all these aspects.

\textbf{Computational Efficiency.}
One of the primary objectives of Sparse-vDiT is to achieve significant inference acceleration without compromising output quality. On CogVideoX1.5, it reduces computational cost from 147.87 to 70.69 PFLOPS (52.2\% reduction), and on HunyuanVideo, from 612.37 to 257.09 PFLOPS (57.9\%). These are the lowest among all compared methods, demonstrating the effectiveness of our sparsity strategy.
In real-world latency, Sparse-vDiT consistently outperforms all baselines, reducing inference time from 901 seconds to 511 seconds on CogVideoX1.5 and from 3166 seconds to 1715 seconds on HunyuanVideo. These improvements are critical for time-sensitive applications.
In terms of speedup, Sparse-vDiT achieves the highest ratios: 1.76$\times$ on CogVideoX1.5 and 1.85$\times$ on HunyuanVideo, surpassing all baseline methods. These results highlight the practical advantages of our sparsity.

Overall, Sparse-vDiT achieves an optimal trade-off between generation quality and efficiency, setting a new state-of-the-art for accelerated vDiT. These results confirm that Sparse-vDiT is not only a theoretically elegant solution but also a highly practical one, enabling scalable deployment of vDiT in latency-sensitive applications.

\begin{table}[]
\centering
\small
\caption{Ablation study on the effects of hyperparameters $\lambda$ and $\epsilon$ in Sparse-vDiT.}
\resizebox{0.85\linewidth}{!}{
\begin{tabular}{cccccccc}
\hline
 \multicolumn{2}{c|}{Hyperparameter}    & \textbf{SSIM 
$\uparrow$}& \textbf{PSNR $\uparrow$}& \textbf{LPIPS $\downarrow$}& \textbf{ImageQual $\uparrow$}& \textbf{SubConsist $\uparrow$}&\textbf{Speedup $\uparrow$}\\ \hline
 \multirow{4}*{$\lambda$}&\multicolumn{1}{c|}{0}   & 0.8182 & 24.0864 & 0.1501 & 63.37\%& 92.61\%&1.74$\times$\\
 &\multicolumn{1}{c|}{0.1} & 0.8180 & 24.0558 & 0.1503 & 63.35\%& 92.62\%&1.73$\times$\\
 &\multicolumn{1}{c|}{0.5} & 0.8212 & 24.1311 & 0.1477 & 63.45\%& 92.66\%&1.76$\times$\\
 &\multicolumn{1}{c|}{1}   & 0.8203 & 24.0946 & 0.1479 & 63.37\%& 92.58\%&1.73$\times$\\ \hline
 \multirow{5}*{$\epsilon$}&\multicolumn{1}{c|}{0.5}   & 0.8512& 25.4929& 0.1219& 63.26\%& 92.60\%&1.68$\times$\\
 &\multicolumn{1}{c|}{1} & 0.8212& 24.1311& 0.1477& 63.45\%& 92.66\%&1.76$\times$\\
 &\multicolumn{1}{c|}{3} & 0.7883& 22.7048& 0.1785& 63.34\%& 92.66\%&1.81$\times$\\
 &\multicolumn{1}{c|}{5}   & 0.7716& 22.0171& 0.1947& 63.27\%& 92.45\%&1.87$\times$\\ 
 &\multicolumn{1}{c|}{10} & 0.7399& 20.8411& 0.2231& 63.30\%& 92.49\%&1.91$\times$\\
\hline
 
\end{tabular}
}
\vspace{-10pt}
\label{tab:ablation}
\end{table}

\vspace{-5pt}
\subsection{Ablation}
\vspace{-5pt}
There are two hyperparameters in Sparse-vDiT, $\lambda$ and $\epsilon$. The parameter $\lambda$ controls the trade-off between efficiency loss and quality loss, while $\epsilon$ regulates the overall sparsity of the vDiT. This section analyzes their impact through experiments on CogVideoX1.5.

\textbf{Quality-Efficiency trade-off.} With $\epsilon$ fixed at its optimal value of 1, we vary $\lambda$ across 0, 0.1, 0.5, and 1. Results are reported in Table~\ref{tab:ablation}. Comparisons across metrics show that both $\lambda$ = 0.5 and $\lambda$ = 1 yield strong generation quality. However, $\lambda$=1 is less efficient. Thus, $\lambda$ = 0.5 offers a better trade-off between generation quality and efficiency, and is used as the default configuration in Table~\ref{tab:main_results}.

\textbf{Performance under different levels of sparsity.} Fixing $\lambda$ at 0.5, we evaluate $\epsilon$ values of 0.5, 1, 3, 5, and 10. Table~\ref{tab:ablation} illustrates that increasing $\epsilon$ leads to greater sparsity, resulting in higher acceleration. For instance, $\epsilon$ = 10 achieves a speedup of 1.91$\times$. However, higher sparsity can impair the quality of generation, as reflected in performance metrics. Notably, at $\epsilon$ = 5, Sparse-vDiT achieves a 1.87$\times$ speedup while still outperforming the SVG baseline (1.64$\times$ speedup). In practice, $\epsilon$ can be adjusted to achieve the desired balance between quality and efficiency.

\section{Conclusion and Limitation}
We propose Sparse-vDiT, an efficient inference method for vDiT based on structured sparsity. It combines predefined sparsity patterns with an offline diffusion-guided search to assign the most suitable configuration to each attention head. Experiments on CogVideo and HunyuanVideo demonstrate theoretical speedups of 2.09$\times$ and 2.38$\times$, and actual speedups of 1.76$\times$ and 1.85$\times$, respectively. Despite the acceleration, video quality remains comparable to that of the original models, with PSNR values of 24.13 and 27.09. These results highlight Sparse-vDiT’s ability to balance efficiency and generation quality, establishing a new state-of-the-art for sparsity-based vDiT acceleration. 

\textbf{Limitation}: In our framework, the sparse kernel for attention is predefined. 
However, in practice, the predefined sparsity level may not fully align with the actual sparsity of the attention maps, potentially leading to under- or over-sparsification. We believe that enabling adaptive sparsity adjustment based on the characteristics of the attention maps, or establishing a more principled approach to sparsity design, could further enhance both sparsification effectiveness and generative performance.

  \bibliography{egbib}
  \bibliographystyle{plain}


\newpage
\appendix

\textbf{{\Large Appendix for \methodshort{}}}

\section{Algorithm Implementation}
Figure~\ref{fig:main} presents the overall process of the offline sparse diffusion search algorithm, with implementation details provided in the accompanying pseudocode. By optimizing across layers and heads, the algorithm selects attention patterns for each head in vDiT. These optimized patterns are subsequently used to accelerate inference.




\begin{algorithm}[h]
\caption{Offline Sparse Diffusion Search} 
\label{algo:algo} 
\KwIn{Pretraine vDiT model $P$ ($N$ layers and $H$ heads), hyperparameter $\lambda$ and $\epsilon$, predefined attention pattern $M_i$ and sparsity $S_i$, timestep $T$}
\KwOut{Attention pattern config $f$}
\For{$i ~\textbf{in} 0,...,4$}{
    \Comment{\scriptsize Predefined Attention Kernel.}

    Compile sparse attention $M_i$ accoding to $S_i$
}

\Comment{\scriptsize Offline Sparse Search.}


$f$=[]

$x_T\sim \mathcal{N}(\bf{0},I)$

\For{$t ~\textbf{in} T,...,1$}{
$x^p_t$ = $Preprocess$ $P$($x_t$)

\Comment{\scriptsize vDiT Layers.}
\For{$n ~\textbf{in} 1,...,N$}{

$Q, K, V$ = $Linear$, $RoPE$ and $Norm$ $P$($x_t^p$)

\Comment{\scriptsize Attention Part (Our Optimization Object).}

$loss$ = []

$x^{gt}_t$ = $M_0(Q,K,V)$

$x_t^o$ = $zeros\_ like(x^{gt}_t)$

\For{$i ~\textbf{in} 1,...,4$}{
    $x_t^i$ = $M_i(Q,K,V)$

    $loss.append(MSE(x_t^i,x_t^{gt})+\lambda S_i)$

}

\Comment{\scriptsize Per-Head Optimization.}

\For{$h ~\textbf{in} 1,...,H$}{
\If{$(loss_h > \epsilon).sum \ge 4$}{
    $x_{t,h}^o=x_{t,h}^{gt}$

    $f.append(0)$
}

\Else{
    $i=argmin(loss_h)$
    
    $x_{t,h}^o=x_{t,h}^{i}$

    $f.append(i)$
}
}
\Comment{\scriptsize FFN Part.}
$x_t^p$ = $FFN$ $P(x_t^o)$

}
\Comment{\scriptsize Denoising.}
$x_{t-1}=Solver(x_t, x_t^p)$
}

\algorithmicreturn{$f$}

\end{algorithm}



\section{Performance on more pretrain models}
\begin{table*}[!t]
    \centering
    \small
    \caption{Comparison of video generation quality and efficiency between Sparse-vDiT and the baseline.  All reported efficiency metrics are measured on a single NVIDIA H800 GPU with a batch size of 1.}
    \resizebox{\linewidth}{!}{
    \begin{tabular}{l|ccccccc}
    \toprule
    
    \multirow{3}{*}{\textbf{Method}} & \multicolumn{3}{c}{\textbf{Against Original}}   & \multicolumn{2}{c}{\textbf{VBench Score}} & \multirow{3}{*}{\textbf{Latency} ($\downarrow$)} & \multirow{3}{*}{\textbf{Speedup} ($\uparrow$)} \\
     \cmidrule(lr){2-4} \cmidrule(lr){5-6}
     & \textbf{SSIM} ($\uparrow$) & \textbf{PSNR} ($\uparrow$) & \textbf{LPIPS} ($\downarrow$) & \textbf{ImageQual } ($\uparrow$) & \textbf{SubConsist} ($\uparrow$) &   &      \\
    \midrule
    Wan2.1& -   & -  & -   & 67.61\%& 91.95\%& 1935s& 1.00$\times$\\
    SVG& 0.78& 21.96& 0.18& 67.18\%& 91.27\%& 1298s& 1.49$\times$\\
     \rowcolor[HTML]{EFEFEF} 
    Sparse-vDiT (Ours)& 0.80& 22.59& 0.16& 67.35\%& 91.39\%& 1228s& 1.58$\times$\\
 \rowcolor[HTML]{EFEFEF} 
    Sparse-vDiT + FP8 (Ours)& 0.79& 22.39& 0.16& 67.22\%& 91.29\%& 1089s&1.78$\times$\\
    \bottomrule
    \end{tabular}
    }
    \label{tab:wan_results}
\end{table*}

Recent models with a Self-Attn and Cross-Attn structure, such as wan2.1, have also demonstrated strong performance. To further assess Sparse-vDiT, we evaluate it under this architecture as well. As shown in Table~\ref{tab:wan_results}, 
our Sparse-vDiT framework introduces a sparse video Diffusion Transformer that achieves a 1.78× speedup (1089ms vs. 1298ms) over SVG while maintaining superior perceptual quality (SSIM: 0.80, LPIPS: 0.16), leveraging structured sparsity and FP8 quantization to reduce latency by 17\% with negligible quality degradation (<0.5\% drop on VBench), outperforming Wan2.1 and SVG across all metrics (+0.0204 SSIM) and demonstrating hardware-efficient scalability on H800 GPUs with near-linear acceleration, bridging the gap between theoretical sparsity and practical deployment in diffusion-based video generation.


\section{More visual results}
\begin{figure*}[!t]
    \centering
    \includegraphics[width=0.99 \linewidth]{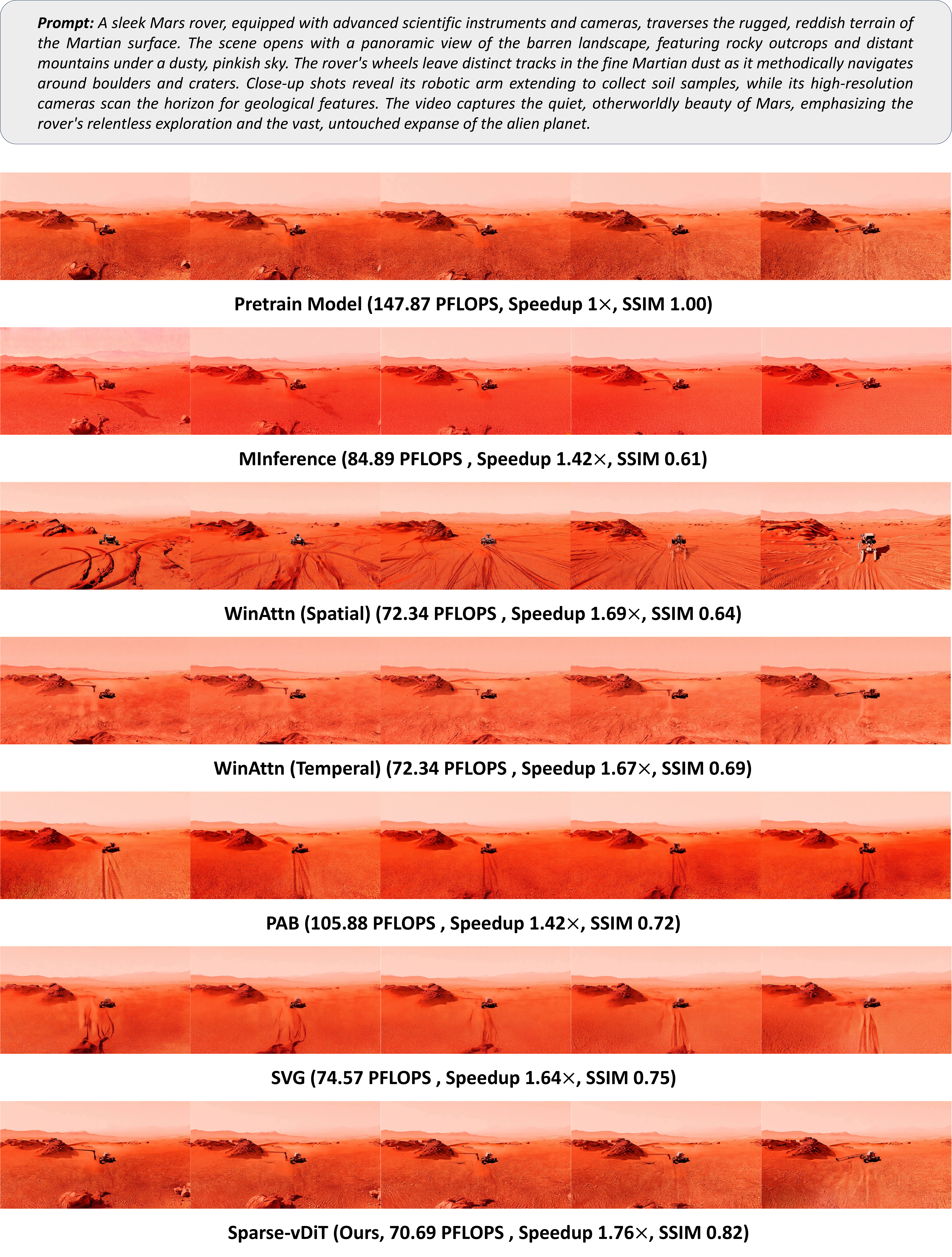}
    \caption{More visual comparison between the proposed Sparse-vDiT and the baseline method. Our method maximizes computational speedup while maintaining high fidelity to the pretrain model.}
    \label{fig:appendix_visual2}
\end{figure*}

\begin{figure*}[!t]
    \centering
    \includegraphics[width=0.99 \linewidth]{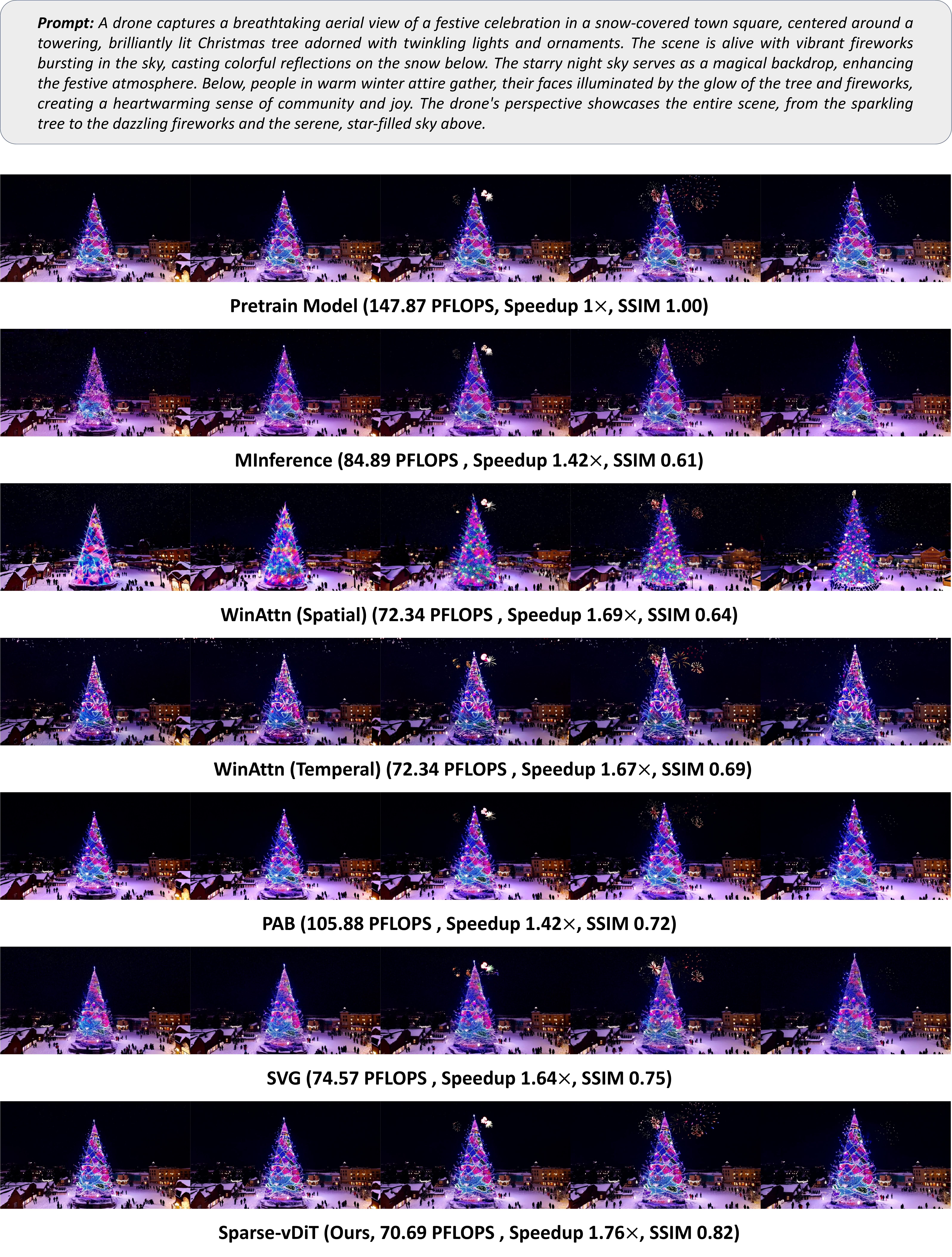}
    \caption{More visual comparison between the proposed Sparse-vDiT and the baseline method. Our method maximizes computational speedup while maintaining high fidelity to the pretrain model.}
    \label{fig:appendix_visual1}
\end{figure*}

\begin{figure*}[!t]
    \centering
    \includegraphics[width=0.99 \linewidth]{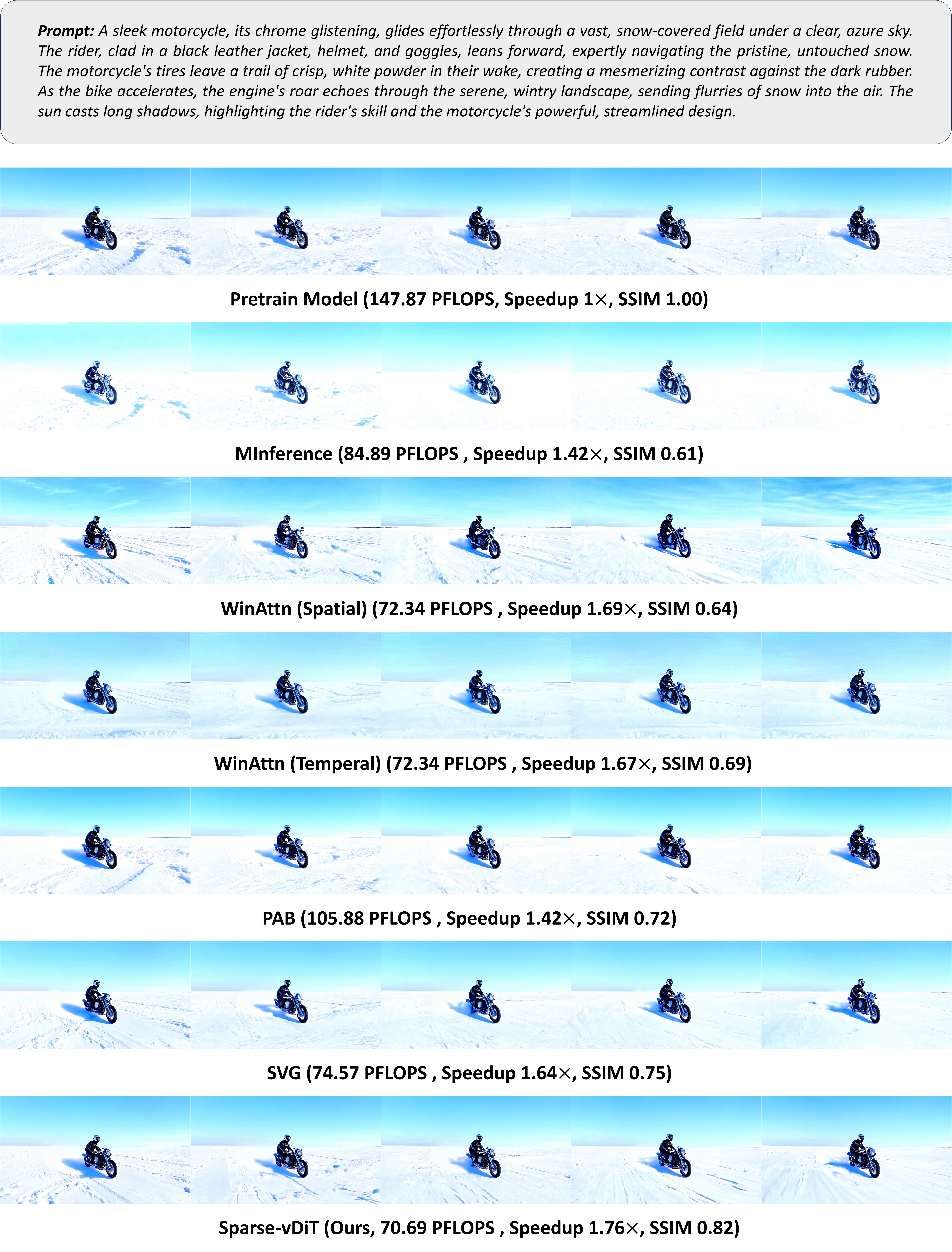}
    \caption{More visual comparison between the proposed Sparse-vDiT and the baseline method. Our method maximizes computational speedup while maintaining high fidelity to the pretrain model.}
    \label{fig:appendix_visual4}
\end{figure*}

\begin{figure*}[!t]
    \centering
    \includegraphics[width=0.99 \linewidth]{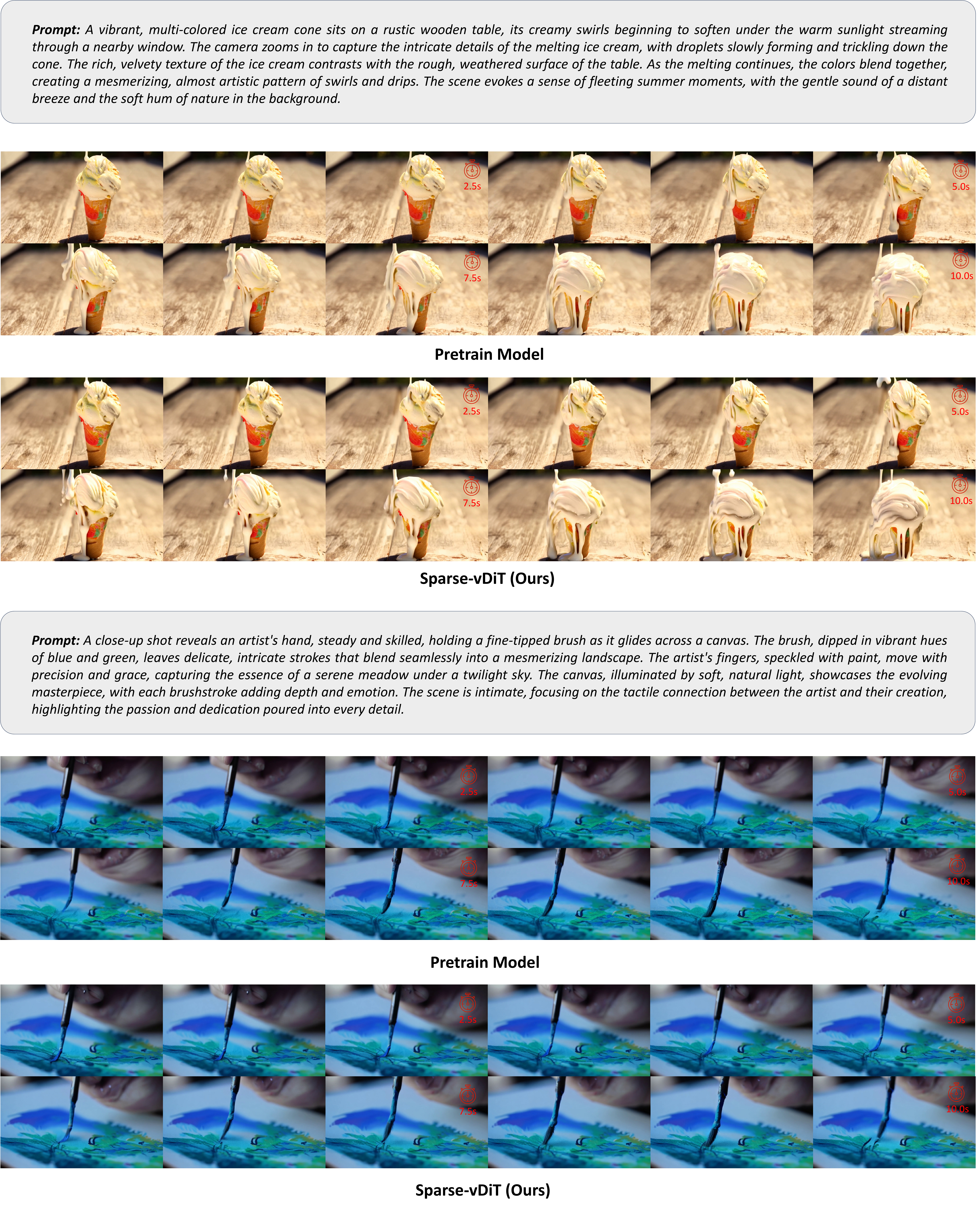}
    \caption{More visual comparison between the proposed Sparse-vDiT and the pretrain model. Beyond demonstrating superior performance in frame generation quality, our method exhibits robust capabilities in maintaining inter-frame consistency.}
    \label{fig:appendix_visual3}
\end{figure*}

\begin{figure*}[!t]
    \centering
    \includegraphics[width=0.99 \linewidth]{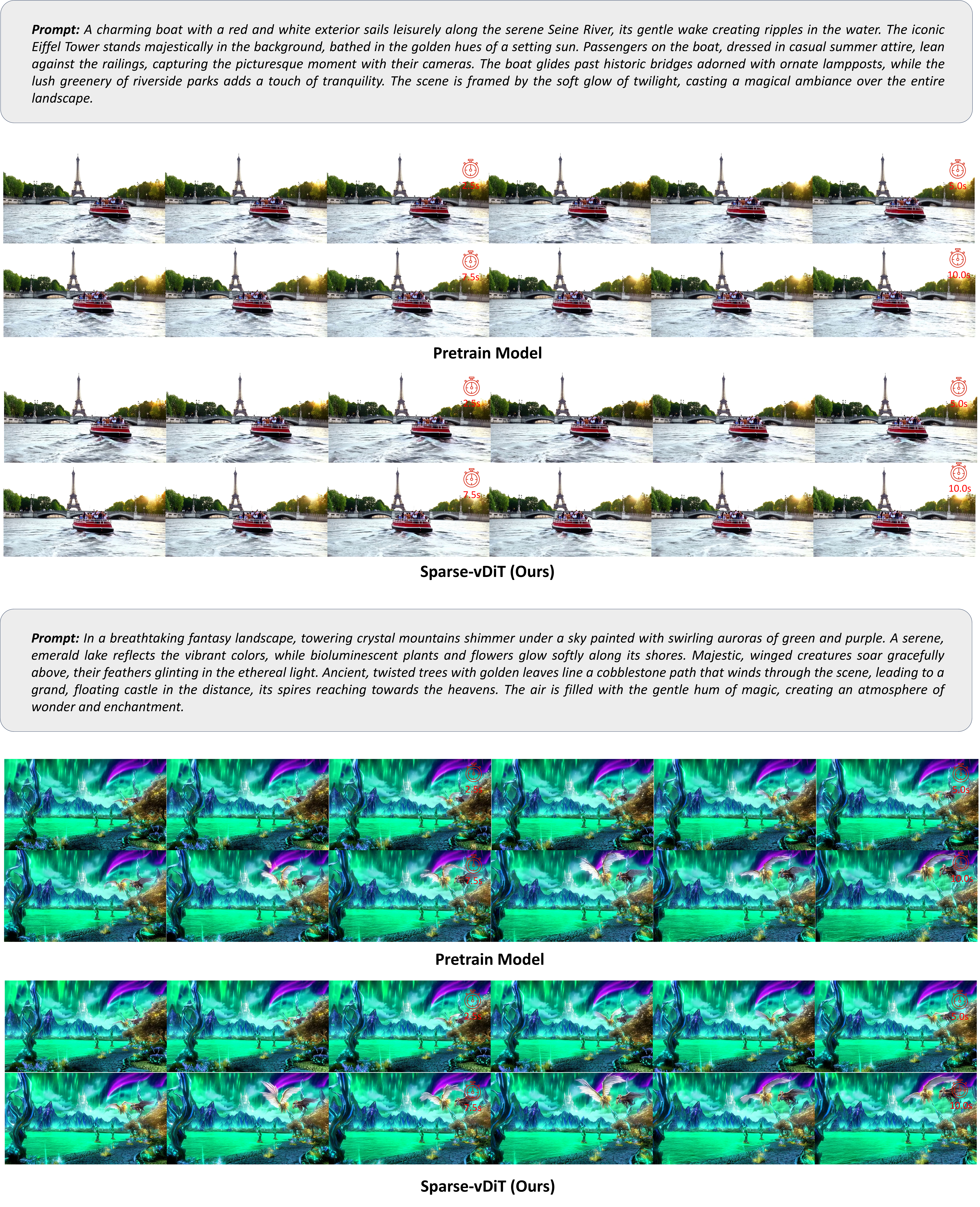}
    \caption{More visual comparison between the proposed Sparse-vDiT and the pretrain model. Beyond demonstrating superior performance in frame generation quality, our method exhibits robust capabilities in maintaining inter-frame consistency.}
    \label{fig:appendix_visual5}
\end{figure*}

Due to space constraints, the main manuscript only compares visualization results for a limited set of baseline methods. Here, we present additional visualizations. Figures~\ref{fig:appendix_visual1}, Figure ~\ref{fig:appendix_visual2}, and Figure~\ref{fig:appendix_visual4} compare our method against all baselines. The WinAttn method exhibits significant contour shifts, while SVG shows smaller deviations. PAB and MInference suffer from frame smoothing and blurring. In contrast, our method preserves the overall contour consistent with the pretrained model and achieves the highest acceleration ratio to date, effectively balancing generation speed and quality. Beyond individual frame quality, frame-to-frame consistency is visualized in Figure~\ref{fig:appendix_visual3} and Figure~\ref{fig:appendix_visual5}. Sparse-vDiT closely matches the pretrained model’s temporal consistency, indicating strong frame coherence. The visualization results start on the next page.

\end{document}